%% file: arxiv1.tex
\documentclass[nofootinbib,onecolumn,prx, superscriptaddress,longbibliography,margin=1mm 11pt]{revtex4-2} 
\usepackage{enumitem}
\usepackage{algorithm}
\usepackage{algorithmicx}
\usepackage{lineno}
\usepackage{booktabs}
\usepackage{float}
\usepackage{xcolor}
\usepackage{tcolorbox}
\usepackage{enumitem}
\usepackage{newfloat}
\usepackage{listings}

\input{math_commands.tex}

\usepackage[normalem]{ulem} % 放在导言区
\usepackage{url}
\usepackage{amsmath}
\usepackage{graphicx}
\usepackage{booktabs}
\usepackage{multirow}
\usepackage{makecell}
\usepackage{amsthm}
\usepackage{xcolor}
\usepackage{tcolorbox} %cellbox
\usepackage{multirow} 
\usepackage{array}
\usepackage{graphicx}
\usepackage{tikz}
\usepackage{callouts}
\usepackage{bbm}
\usepackage{caption,subcaption}
\usepackage{mathtools}
\usepackage{amsmath,bm}
\newcommand{\bd}[1]{\bm{#1}}   % 用法：\bd{s}, \bd{\alpha}, \bd{\Sigma}
\usepackage{booktabs}        % \toprule \midrule \bottomrule
\usepackage{siunitx}         % S 列对齐数值
\usepackage[table]{xcolor}   % \rowcolor，需加 [table] 选项
\usepackage{threeparttable}  % threeparttable 环境
\usepackage{adjustbox}       % adjustbox/max width
\usepackage{subcaption}      % 子表(subtable)；依赖 caption 包，subcaption 会处理

\usepackage[textfont=small,labelfont=small]{caption}
\usepackage{wrapfig}
\usepackage{amsfonts}
\usepackage{lineno}
\usepackage[english]{babel}

\makeatletter
\adddialect\l@en\l@english   % 把语言代码 'en' 定义成 english 的一个方言
\makeatother

\usepackage{ragged2e}
\usepackage{graphicx}
\usepackage{bm}

\usepackage{environ}
\usepackage{appendix}
\usepackage{physics}
\usepackage{rotating,booktabs,array,amsmath,amssymb}
\newcolumntype{C}{>{$\displaystyle}c<{$}}
\usepackage{caption}
\usepackage{subcaption}
\usepackage[nopar]{lipsum}
\usepackage{enumitem}
\setlist[enumerate,1]{label=(\arabic*),ref=\arabic*}
\makeatletter
\usepackage{amsthm}
\usepackage{amssymb}
\usepackage{amsfonts}
\usepackage{fancyhdr}
\usepackage{slashed}
\usepackage{bbm}
\usepackage{latexsym,epsfig,bbm}
\usepackage[bottom]{footmisc}
\usepackage[colorlinks=true,urlcolor=blue,citecolor=blue]{hyperref}
\definecolor{blue}{rgb}{0,0.2,1}
\definecolor{LightCyan}{RGB}{235,248,255}

\definecolor{red}{rgb}{0.9,0,0}

\newtheorem{theorem}{Theorem}

\newtheorem{definition}[theorem]{Definition}

\definecolor{aqua}{rgb}{0.00,0.67,0.80}

\newcommand{\inc}[1]{\textcolor{red!50!black}{$\uparrow$~#1\%}}
\newcommand{\dec}[1]{\textcolor{green!50!black}{$\downarrow$~#1\%}}
\newcommand{\same}{\textcolor{black!55}{$\leftrightarrow$~0\%}} % 持平

\begin{document}
\title{Improved Offline Reinforcement Learning via Quantum Metric Encoding}
\author{Outongyi Lv}
\email{harry\_lv@sjtu.edu.cn}
\thanks{These authors contributed equally to this work and are listed alphabetically.}
% \affiliation{Institute of Natural Sciences, Shanghai Jiao Tong University, Shanghai 200240, China}
\affiliation{School of Mathematical Sciences, Shanghai Jiao Tong University, Shanghai 200240, China}

\author{Yewei Yuan}
\email{yuanyw@sjtu.edu.cn}
\thanks{These authors contributed equally to this work and are listed alphabetically.}
\affiliation{Global College, Shanghai Jiao Tong University, Shanghai 200240, China}

\author{Nana Liu}
\email{nana.liu@quantumlah.org}
% \email{nana.liu@quantumlah.org}
\affiliation{School of Mathematical Sciences, Shanghai Jiao Tong University, Shanghai 200240, China}
\affiliation{Institute of Natural Sciences, Shanghai Jiao Tong University, Shanghai 200240, China}
% \affiliation{Ministry of Education Key Laboratory in Scientific and Engineering Computing, Shanghai Jiao Tong University, Shanghai 200240, China}
% \affiliation{Shanghai Artificial Intelligence Laboratory, Shanghai, China}
\affiliation{Global College, Shanghai Jiao Tong University, Shanghai 200240, China}

\begin{abstract}
  Reinforcement learning (RL) with limited samples is common in real-world applications.    However, offline RL performance under this constraint is often suboptimal. We consider an alternative approach to dealing with limited samples by introducing the Quantum Metric Encoder (QME). In this methodology, instead of applying the RL framework directly on the original states and rewards, we embed the states into a more compact and meaningful representation, where the structure of the encoding is inspired by quantum circuits. For classical data, QME is a classically simulable, trainable unitary embedding and thus serves as a \textit{quantum-inspired module}, on a classical device. For quantum data in the form of quantum states, QME can instead be implemented directly on quantum hardware, allowing for training directly without measurement or re-encoding.
   
    We evaluated QME on three datasets, each limited to 100 samples. We use Soft-Actor-Critic (SAC) and Implicit-Q-Learning (IQL), two well-known RL algorithms, to demonstrate the effectiveness of our approach. From the experimental results, we find that training offline RL agents on QME-embedded states with decoded rewards yields significantly better performance than training on the original states and rewards. On average across the three datasets, for maximum reward performance, we achieve a \textbf{116.2\%} improvement for SAC and \textbf{117.6\%} for IQL.
    
    We further investigate the $\Delta$-hyperbolicity of our framework, a geometric property of the state space known to be important for the RL training efficacy.
    The QME-embedded states exhibit low $\Delta$-hyperbolicity, suggesting that the improvement after embedding arises from the modified geometry of the state space induced by QME.  Thus, the low $\Delta$-hyperbolicity and the corresponding effectiveness of QME could provide valuable information for developing efficient offline RL methods under limited-sample conditions.

\end{abstract}

\maketitle

\section{Introduction}
\label{sec:introduction}
Reinforcement learning (RL) has matured into a core machine-learning (ML) paradigm for sequential decision-making under uncertainty. Catalyzed by the breakthrough of AlphaGo \citep{silver2016mastering}, RL now sits at the center of modern AI, with applications spanning large language models \citep{zhu2023principled, wang2023openchat,chen2024self,guo2025deepseek}, high-performance game agents \citep{mnih2013playing, qi2023adaptive}, and clinical decision support \citep{ehrmann2023making, otten2024does, drudi2024reinforcement}. RL is commonly organized into two paradigms: online and offline. The former permits continual interaction with the environment for policy updates; the latter disallows interaction, requiring learning and evaluation solely from a fixed, pre-collected dataset. Since data collection is costly and limited in practice, this makes data-efficient offline learning more aligned with real-world constraints. 

However, compared to online RL, offline RL often underperforms. The performance of offline RL is sensitive to the dataset’s quality and coverage. In the limited-sample regime, insufficient support can exacerbate Q-function overfitting.  Q-function overestimation \citep{kumar2020conservative} and distributional shift \citep{kostrikov2021offline} are considered the main factors contributing to the performance degradation.
Accordingly, recent work advances data-efficient offline learning via conservative value estimation, policy constraints, and dataset-aware regularization \citep{liu2022avoiding,nie2022data,macaluso2024small}. 
Because offline RL disallows further interaction with the environment, the target policy must learn from a fixed dataset generated by other behavior policies. Consequently, most offline RL algorithms are inherently off-policy, i.e., they learn under the state–action distribution induced by policies different from the target policy.
Among off-policy methods, Soft-Actor-Critic (SAC) \citep{haarnoja2018soft2,haarnoja2018soft} and Implicit-Q-Learning (IQL) \citep{kostrikov2021offline} are particularly prominent and effective compared to other methods. SAC introduces an entropy regularization term to maximize policy entropy and avoid premature convergence to suboptimal strategies. IQL performs supervised regression directly on the actions in the dataset to mitigate distribution shift in offline training. However, under limited-sample conditions, their training still fails to demonstrate a clear advantage.

A promising approach to overcoming these offline training challenges is the use of state-reward sample encoding, which allows the learning process to leverage more compact and meaningful representations of states and actions. Building on this idea, Frans et al. \citep{frans2024unsupervised} introduced a framework that decouples the state encoder and reward decoder, training them independently. This method addresses the issue of massive input spaces in offline RL by improving both robustness and performance, suggesting that decoupling state embedding from reward decoding could systematically enhance offline RL training.
% Recently, \citep{frans2024unsupervised} introduced a framework that leverages random sampling and independently trains of a state encoder and a reward decoder, effectively addressing the challenge of massive inputs in offline RL and delivering strong performances. This suggests that decoupling state embedding from reward decoding may systematically improve robustness and performance in offline training. 

In practice, however, data is often limited rather than extremely large. Under such limited-sample conditions, we need to identify methods that can efficiently compress information and capture structural cues that traditional end-to-end pipelines may miss. We therefore turn our attention to methods inspired by quantum encoders as a promising approach. Quantum computing is advancing rapidly and has the potential to revolutionize various fields. By leveraging superposition and entanglement, quantum computers process information fundamentally differently from classical computers, enabling quantum algorithms like Grover's \citep{grover1997quantum} and Shor's \citep{shor1994algorithms} to solve problems more efficiently and achieve quantum advantage. This quantum advantage can also be extended to improve ML \citep{biamonte2017quantum}. For example, quantum feature maps and quantum kernels embed data into high-dimensional Hilbert spaces \citep{havlivcek2019supervised,schuld2019quantum}, which can increase expressive capacity for learning. Within this landscape, data encoded into the quantum state space by Parameterized Quantum Circuits (PQCs) \cite{schuld2021effect} improve ML performance in certain tasks \citep{Perez-Salinas2020-cp}, thus quantum neural networks (QNNs) on PQCs can exhibit higher effective dimension and faster training ability \citep{abbas2021power}.
Meanwhile, some work has shown that trainable quantum circuits can implement metric learning \citep{lloyd2020quantum} and generalize well from limited data relative to classical models \citep{caro2022generalization}. While some of these benefits could arise from mapping to higher dimensional Hilbert space, other benefits could arise from the mathematical structure itself. This means that there are possibilities of quantum-inspired frameworks that utilise the same mathematical structure as sequences of unitary quantum circuits, but which are also implementable on purely classical devices. 

% Building on these potential quantum advantages, quantum RL has rapidly emerged, giving rise to numerous algorithms \citep{dong2008quantum,lan2021variational,kolle2024quantum}, showing improvements in both computational acceleration \citep{zhong2023provably,su2025quantum,chen2024learning} and task performance \citep{ding2022evolutionary,sun2023differentiable}.
Based on the inspiration above, we are naturally led to the following question:

\begin{quote}
\textbf{\textit{``Can we leverage quantum-inspired encoders and decoders to enable effective reward learning and obtain state embeddings suitable for RL with limited samples?"}}
\end{quote}

\begin{figure*}[h!]
    \centering
    \includegraphics[width=\textwidth]{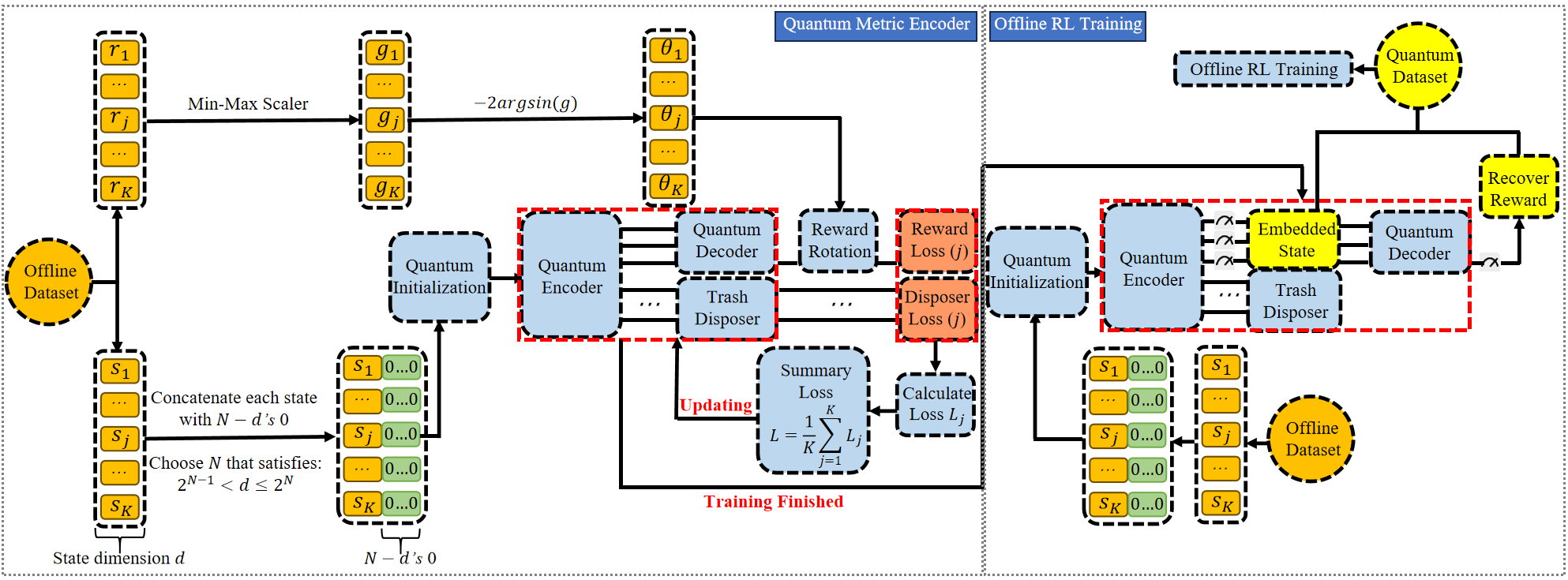}
    \caption{The structure of our model, including the \textbf{QME} and \textbf{Offline RL Training} parts. The structure for QME's details is thoroughly discussed in Section~\ref{sec:method}.}
    \label{fig:Structure}
\end{figure*}

In response to this question, we introduce the Quantum Metric Encoding (QME) framework. QME comprises an encoder–decoder architecture in which both the encoder and decoder are implemented using parameterised unitary circuits, and reward learning is supervised by the decoder. The framework and its downstream pipeline are presented in Fig.~\ref{fig:Structure}. 
QME uses the parameterized unitary operator as an embedding model rather than this necessarily requiring a quantum device. Consequently, the reward-supervised metric learning objective can be simulated and optimized entirely on classical hardware, allowing QME to be viewed as a quantum-inspired algorithm in general settings. Real quantum devices become appealing only when the embedding dimension and circuit width make faithful classical simulation prohibitively costly. Moreover, when the data are quantum-native (e.g., states produced in physical experiments), QME can ingest them directly as input states without measurement and re-encoding.

We then validate QME across three datasets, each limited to 100 samples. 
We integrate the QME-embedded states and decoded rewards into SAC and IQL. We find that, relative to using ground-truth states and rewards, employing QME-embedded states together with rewards yields an average improvement of \textbf{116.2\%} in the maximum evaluation return for SAC and \textbf{117.6\%} for IQL. This indicates that QME can efficiently learn an appropriate reward scoring function from a limited dataset while simultaneously producing embedded states well suited to offline RL. 

In addition, we investigate the role of $\Delta$-hyperbolicity \cite{gromov1987hyperbolic} of the embedded state obtained through QME, which we show to be significantly lower than that of the real state or the normalized real state (in Euclidean space). The magnitude of $\Delta$-hyperbolicity has been shown to be closely linked to RL training performance \cite{gromov1987hyperbolic}. Lower $\Delta$-hyperbolicity is considered to be more compatible with tree-like structures, and it is an important factor in enabling RL to maintain strong performance during testing \citep{cetin2022hyperbolic}. 
The low $\Delta$-hyperbolicity suggests the improvement arises from these QME-embedded states capturing a good metric.
% Building on this, we preliminarily hypothesize that QME- embedded states may hold substantial potential for offline RL training. 
%In summary, our results suggest that QME may offer valuable guidance for future integration of quantum computing and RL.

The structure of the paper is as follows: in Section~\ref{sec:preliminaries} we introduce the foundational concepts of RL (and SAC, IQL algorithms), and parameterised quantum circuits. Section~\ref{sec:method} describes the details of the QME framework and its components. Section~\ref{sec:experiment} presents the experimental validation of QME. Finally, we summarise our results in  Section~\ref{sec:conclusion} and discuss future work.

\section{Preliminaries}
\label{sec:preliminaries}
Before delving into the main concepts in Section~\ref{sec:method}, this section provides preliminary foundational knowledge for the readers.
In Section~\ref{sec:RL's objectives}, we cover the basic principles and objectives of RL. In Section~\ref{sec:SAC_IQL}, we introduce the principles of SAC and IQL. In Section~\ref{sec:PQC}, we introduce the fundamental concepts of PQC.

\subsection{Objectives of reinforcement learning (RL)}
\label{sec:RL's objectives}
In RL, the Markov Decision Process serves as the fundamental framework for describing interactions between the agent and its environment. This process is defined by the tuple $(\gS,\gA,\sP,r,\gamma)$, where $\gS$ and $\gA$ represent the state space and action space. They denote the environment state $\bd{s}$ $(\bd{s} \in \gS)$ observed by the agent and the action $\bd{a}$ $(\bd{a} \in \gA)$ chosen by the agent, respectively. $\sP(\bd{s'}|\bd{s},\bd{a})$ is the transition probability of reaching state $\bd{s'}$ after the agent takes the action $\bd{a}$ in the current state $\bd{s}$. $r(\bd{s}, \bd{a})$ is the reward function that determines the reward based on the current state $\bd{s}$ and action $\bd{a}$. $\gamma$ is the discount factor used to calculate the present value of future rewards, influencing the agent’s trade-off between immediate and future rewards. The target of RL is to determine the action $\bd{a}$ at each state $\bd{s}$ to maximize the cumulative discounted reward under a fixed horizon $T$. We define RL's target as follows:

\begin{definition}
    Policy \(\pi(\bd{a}|\bd{s})\) is a mapping from the state space \(\gS\) to a probability distribution over the action space \(\gA\). It defines the behavior of the agent by specifying the probability of taking action \(\bd{a}\) in state \(\bd{s}\). In RL, the agent seeks to find the optimal policy \(\pi^*(\bd{a}|\bd{s})\) that can maximize the cumulative discounted reward over a fixed horizon $T$, which is commonly used by networks such as Actor-Critic (AC) \citep{konda1999actor}:
    \begin{equation}
    \label{eq:RL-target}
        \pi^*(\bd{a}|\bd{s})=\argmax_{\pi}\left[
        \E_{\bd{a_{t}} \sim      \pi(\bd{a_{t}}|\bd{s_{t}})}\left[\sum_{t=0}^T\gamma^t r(\bd{s_{t}},\bd{a_{t}})\right]\right].
    \end{equation}
\end{definition}

Before introducing SAC and IQL, we first define the \textbf{Q-function} \( Q_{\pi}(\bd{s_t}, \bd{a_t}) \) and \textbf{V-function} \( V_\pi(\bd{s_t}) \). In fact,  $Q_{\pi}$ and $V_{\pi}$ can also be interpreted probabilistically through Bayesian inference \citep{levine2018reinforcement}.

\begin{definition}
    \( Q_{\pi}(\bd{s_t}, \bd{a_t}) \) estimates the expected cumulative reward of taking action \( \bd{a_t} \) in state \( \bd{s_t} \) under a given policy $\pi$. \( V_{\pi}(\bd{s_t}) \) is the expected cumulative reward of being in state \( \bd{s_t} \) under a given policy $\pi$. $Q_{\pi}$ and $V_{\pi}$ are expressed as follows:
    $$
    Q_\pi(\bd{s_k},\bd{a_k}) = \E_{\bd{s_{k+1}}\sim \sP(\cdot|\bd{s_k},\bd{a_k}),\bd{a_{k+1}\sim\pi(\cdot|\bd{s_{k+1}})}}
    \left[\sum_{t=k}^T\gamma^{t-k} r(\bd{s_{t}},\bd{a_{t}})\right],
    $$
    $$
    V_{\pi}(\bd{s_k}) = \E_{\bd{a_k}}\left[ Q_\pi(\bd{s_k},\bd{a_k})\right].
    $$
    Furthermore, the relationship between $Q_{\pi}$ and $V_{\pi}$ can be expressed as:
    \begin{equation}
    \label{eq: part1}
    Q_\pi(\bd{s_t}, \bd{a_t}) = r(\bd{s_t}, \bd{a_t}) + \gamma \mathbb{E}_{\bd{s_{t+1}} \sim \sP(\cdot|\bd{s_t}, \bd{a_t})} [V_\pi(\bd{s_{t+1}})]
    \end{equation}
\end{definition}

In the subsequent sections, we omit the policy subscript $\pi$ on $Q_\pi$ and $V_\pi$ functions, referring to them simply as $Q$ and $V$.

\subsection{Soft-Actor-Critic (SAC) and Implicit-Q-Learning (IQL) }
\label{sec:SAC_IQL}
Soft-Actor-Critic (SAC)  and Implicit-Q-Learning (IQL)  are two particularly prominent and effective off-policy methods compared to other methods.  
First, SAC posits that the objective in (\ref{eq:RL-target}) should employ a soft update mechanism under the regularization strength $\zeta$, which is achieved by incorporating entropy terms into the policy optimization process to encourage exploration, thus facilitating a more robust learning process. SAC's target is to find the policy $\pi(\bd{a}|\bd{s})$ which can maximize:
\begin{equation}
\label{eq:SAC-target}
  \E_{\bd{a_t} \sim \pi(\bd{a_t}|\bd{s_t})}\left[\sum_{t=0}^T\gamma^t \left(r(\bd{s_t},\bd{a_t})+\zeta S(\pi(\cdot|\bd{s_t})) \right)\right],
\end{equation}

where $S$ is the entropy of a policy \(\pi\) at state \( \bd{s_t} \): $S(\pi(\cdot|\bd{s_t})) = -\sum_{a_t\in \gA}\pi(\bd{a_t}|\bd{s_t})\log(\pi(\bd{a_t}|\bd{s_t}))$. This problem can be solved by the soft Bellman iteration \citep{haarnoja2018soft2}, where $k$ denotes the iteration step:
\begin{equation}
\label{eq:SAC-Q-iteration}
Q^{k+1}(\bd{s_t},\bd{a_t})=r(\bd{s_t},\bd{a_t})+\E_{\bd{s_{t+1}}\sim \sP(\cdot|\bd{s_t},\bd{a_t}),\bd{a_{t+1}}\sim \pi(\cdot|\bd{s_{t+1}})}\biggl[Q^{k}(\bd{s_{t+1}},\bd{a_{t+1}})-\zeta \log(\pi(\bd{a_{t+1}}|\bd{s_{t+1}}))\biggr]
\end{equation}

with the optimal policy $\pi^{SAC}(\bd{a_{t}}|\bd{s_{t}})$ given by:
\begin{equation}
\label{eq: optimal Policy SAC}
  \pi^{SAC}(\bd{a_{t+1}}|\bd{s_{t+1}})=\argmax_{\pi}(\E_{\bd{s_{t+1}}\sim \sP(\cdot|\bd{s_t},\bd{a_t}),\bd{a_{t+1}}\sim \pi(\cdot|\bd{s_{t+1}})}\left[Q(\bd{s_{t+1}},\bd{a_{t+1}})-\zeta \log(\pi(\bd{a_{t+1}}|\bd{s_{t+1}}))\right]).
\end{equation}

Since $\sum_{\bd{a_t}} \pi^{SAC}(\bd{a_t}|\bd{s_t})=1$, applying the Lagrange multiplier method \citep{bertsekas2014constrained}, we derive the following explicit expression from (\ref{eq: optimal Policy SAC}):
\begin{equation}
\label{eq: optimal result Policy SAC}
    \pi^{SAC}(\bd{a_t}|\bd{s_t})=\frac{e^{Q(\bd{s_t},\bd{a_t})/\zeta}}{\sum_{\bd{a_t}} e^{Q(\bd{s_t},\bd{a_t})/\zeta}}.
\end{equation}
But it is quite challenging to estimate (\ref{eq:SAC-Q-iteration}) directly. Therefore, the SAC approach considers separating the latter part to be predicted using the function $V$ independently. It separates (\ref{eq:SAC-Q-iteration}) into the following two parts (\ref{eq: part1}, \ref{eq: part2}):

\begin{equation}
\label{eq: part2}
    V(\bd{s_{t+1}})=\E_{\bd{a_{t+1}}\sim \pi(\cdot|\bd{s_{t+1}})}\biggl[Q(\bd{s_{t+1}},\bd{a_{t+1}})-\zeta \log(\pi(\bd{a_{t+1}}|\bd{s_{t+1}}))\biggr].
\end{equation}

Consequently, training SAC requires three networks: \( Q_\phi \), \( V_\eta \), and \( \pi_\tau \), parameterized by \( \phi \), \( \eta \), and \( \tau \), respectively, which are updated using the equations (\ref{eq: part1}, \ref{eq: part2}, and \ref{eq: optimal result Policy SAC}) for each function.
\\

In offline RL, since there is no interaction with the environment and only a dataset $\gD$ is available, where $\gD$ consists of many tuples of the form $(\bd{s_t}, \bd{a_t}, \bd{s_{t+1}}, \bd{a_{t+1}})$, the goal is to learn $Q_\phi$ that can minimize:
\begin{equation}
\label{eq: offline target}
    L(\phi) = \E_{(\bd{s_t}, \bd{a_t}, \bd{s_{t+1}})\sim \gD}[(r(\bd{s_t},\bd{a_t})+\gamma \max_{\bd{a_{t+1}}} Q_{\phi^T}(\bd{s_{t+1}},\bd{a_{t+1}})-Q_\phi(\bd{s_t},\bd{a_t}))^2],
\end{equation}
where $Q_{\phi^T}$ is the target network updated by Polyak averaging \citep{lillicrap2015continuous}. Polyak averaging updates a target network by taking a weighted average of the current parameters and the previous parameters. This method helps stabilize learning by preventing large oscillations in the target network.
Since offline RL cannot interact with the environment, the max operator tends to overestimate the $Q$ values for states that are out of the policy distribution (OOD). This error amplifies as the training progresses. In order to solve the OOD problem, IQL restricts the actions on the behavior policy $\pi_{\beta}$:
\begin{equation}
\label{eq: iql target1}
    L(\phi) = \E_{(\bd{s_t}, \bd{a_t}, \bd{s_{t+1}})\sim \gD}[(r(\bd{s_t},\bd{a_t})+\gamma \max_{\bd{a_{t+1}},s.t.\pi_{\beta(\bd{a_{t+1}}|\bd{s_{t+1}})}>0} Q_{\phi^T}(\bd{s_{t+1}},\bd{a_{t+1}})-Q_\phi(\bd{s_t},\bd{a_t}))^2].
\end{equation}

The other method, IQL, defines its loss function using expected regression, where the target of IQL is specifically given by:
\begin{equation}
\label{eq: iql target2}
    L(\phi) = \E_{(\bd{s_t}, \bd{a_t}, \bd{s_{t+1}},\bd{a_{t+1}})\sim \gD}[L_2^\tau(r(\bd{s_t},\bd{a_t})+\gamma Q_{\phi^T}(\bd{s_{t+1}},\bd{a_{t+1}})-Q_\phi(\bd{s_t},\bd{a_t}))],
\end{equation}
where
$$
L_2^\tau(u) = |\tau-1(u<0)|u^2. 
$$

In fact, IQL decomposes (\ref{eq: iql target2}) into separate learning tasks for $V_\eta$ and $Q_\phi$:

\begin{equation}
\label{eq: iql V}
    L_V(\eta)=\E_{(\bd{s_t},\bd{a_t})\sim \gD}[L_2^\tau(Q_\phi(\bd{s_t},\bd{a_t})-V_\eta(\bd{s_t}))].
\end{equation}

\begin{equation}
\label{eq: iql Q}
    L_Q(\phi)=\E_{(\bd{s_t},\bd{a_t},\bd{s_{t+1}})\sim \gD}[(r(\bd{s_t},\bd{a_t})+\gamma V_\eta(\bd{s_{t+1}})-Q_\phi(\bd{s_t},\bd{a_t}))^2].
\end{equation}

In addition, IQL uses Advantage-Weighted Regression (AWR) \citep{peng2019advantage}, which is a method used to update the policy by weighting the regression error based on the advantage function, to update the policy $\pi_\tau$ as (\ref{eq: iql awr}). The advantage function measures the relative benefit of an action compared to the average action, helping the agent focus on improving actions with higher returns. 
\begin{equation}
\label{eq: iql awr}
    L_\pi(\tau) = \E_{(\bd{s_t},\bd{a_t})\sim \gD}[e^{\beta(Q_{\phi^T}(\bd{s_t},\bd{a_t})-V_\eta(\bd{s_t}))}\cdot \log[\pi_\tau(\bd{a_t}|\bd{s_t})]],
\end{equation}
where $\beta \in [0,\infty)$ is the inverse temperature parameter influencing the trade-off between exploration and exploitation. Higher values of \( \beta \) lead to more deterministic policies, while lower values encourage more exploration.
Consequently, IQL uses (\ref{eq: iql V}, \ref{eq: iql Q} and \ref{eq: iql awr}) to learn $V_\eta$ $Q_\phi$ and $\pi_\tau$.

\subsection{Parameterised quantum gates and circuits}
\label{sec:PQC}
In quantum computing, a quantum state is a vector \( |\psi\rangle \in \mathcal{H} \), where \( \mathcal{H} \) is a complex Hilbert space describing a quantum system. For a single qubit \( |\psi\rangle = \alpha |0\rangle + \beta |1\rangle \) with \( \alpha, \beta \in \mathbb{C} \), where \( |0\rangle \) and \( |1\rangle \) denote the computational basis states of a single qubit, and \( \mathbb{C} \) denotes the field of complex numbers. For \( n \) qubits, \(|\psi\rangle = \sum_i c_i |i\rangle \), where \( \{ |i\rangle \} \) is an orthonormal basis and \( c_i \in \mathbb{C} \). The state is normalized, i.e., \( \langle \psi | \psi \rangle = \sum_i |c_i|^2 = 1 \), ensuring total probability of measurement outcomes equals one. 
A quantum gate is a unitary operator \( U \) that acts on a quantum state \( |\psi\rangle \) and transforms it into another quantum state \( U|\psi\rangle \). Since \(U\) is unitary, i.e. \(U^\dagger U=I\), it preserves norms and implements reversible dynamics in quantum circuits. A parameterized quantum gate is a quantum gate that depends on one or more parameters, denoted by \( \theta \), such as \( U(\theta) \), allowing the operation to be optimized during computation.

% Quantum computing operates using qubits, represented as state vectors in a complex Hilbert space, capable of superpositions of basis states to enable parallel computation. Quantum gates, implemented as unitary operators, reversibly transform these states while preserving their normalization.
Parameterized quantum circuits consists of a sequence of parameterized quantum gates. Variational Quantum Algorithms (VQAs) \citep{Cerezo2021-zg} based on PQC are investigated for potential quantum advantages on noisy intermediate-scale quantum (NISQ) computers.
The configuration of a PQC can be expressed mathematically as:
\begin{equation}
    U(\boldsymbol{\theta}) = U_n(\theta_n) \ldots U_2(\theta_2) U_1(\theta_1),
\end{equation}
where $U_i(\theta_i)$ are the parameterized quantum gates, and $\boldsymbol{\theta}=(\theta_1,\theta_2,\dots,\theta_n)$ represents the vector of trainable parameters updated during hybrid classical–quantum optimization. These parameters are analogous to the weights in classical neural networks and are optimized to minimize a task-specific cost (loss) function. In quantum ML, a quantum neural network (QNN) typically consists of a data-encoding unitary with data-dependent parameters followed by a trainable PQC. 
The cost (loss) function is often constructed from the
measurement statistics of the circuit’s output state $|\psi_{\boldsymbol{\theta}}\rangle$, most commonly
the expectation value $\langle \hat{O} \rangle=\big\langle \psi_{\boldsymbol{\theta}} \big| \hat{O} \big| \psi_{\boldsymbol{\theta}} \big\rangle$ of a chosen observable $\hat{O}$—a Hermitian operator whose eigenvalues are
the possible measurement outcomes.
In practice, $\langle \hat{O} \rangle$ is estimated from repeated measurements (shots), and the resulting scalar is supplied to a classical optimizer to update $\boldsymbol{\theta}$.
\\

PQCs can also implement a quantum autoencoder (QAE)~\citep{romero2017quantum} to compress quantum data. A QAE comprises an encoder and a decoder. The encoder circuit compresses the input quantum state into a lower-dimensional subspace while concentrating redundancy into $n_{trash}$ ``trash'' qubits, which ideally are reset to $\ket{0}^{\otimes n_{trash}}$. And the decoder reconstructs the original state from the compressed representation. Let $|\psi\rangle$ be the initial quantum state of the system. The encoder aims to transform this state into a lower-dimensional state $|\phi\rangle$, which can be expressed as:
\begin{equation}\label{eq:qae-factorization}
  |\phi\rangle \otimes |0\rangle ^ {\otimes n_{trash}} = U_{\text{e}}(\theta_{\text{e}}) |\psi\rangle,
\end{equation}
where $U_{\text{e}}(\theta_{\text{e}})$ represents the unitary operation performed by the encoder, parameterized by $\theta_{\text{e}}$. And the last $n_{trash}$ qubits are regarded as trash qubits. Accordingly, the encoder is trained by a loss that maximizes the fidelity of the trash qubits with $|0\rangle ^ {\otimes n_{trash}}$, equivalently, minimizes the population outside that subspace:
% A common choice is the expected Hamming weight of the trash register relative to $|0\rangle ^ {\otimes n_{trash}}$, namely
\begin{align}
\label{eq: qae}
  \mathcal{L}(\theta_{\mathrm{e}})&=\frac{1}{2k}\sum_{i=1}^{n_{trash}}\!\bigl(1-\langle Z_i\rangle\bigr)
\\
   &=\frac{1}{n_{trash}}\sum_{i=1}^{n_{trash}}\Pr\bigl(\text{trash qubit }i=\ket{1}\bigr).
\end{align}

where $\langle Z_i\rangle$ denotes the expectation value of the Pauli-$Z$ observable on the $i$-th trash qubit locally. Equivalently, $\frac{1-\langle Z_i\rangle}{2}$ is the probability of measuring $|1\rangle$ on the $i$-th qubit. Minimizing $\mathcal{L}$ encourages the encoder to concentrate redundancy entirely onto the trash register and to prepare it in $|0\rangle ^{\otimes n_{trash}}$, thereby approaching the target factorization in
(\ref{eq:qae-factorization}). Once this condition is (approximately)
met, applying the decoder $U_{\text{e}}^{\dagger}(\theta_{\text{e}})$
to the compressed state recovers the input:
\begin{equation}
  U_{\text{e}}^{\dagger}(\theta_{\text{e}})
  \bigl(\ket{\phi}\otimes\ket{0}^{\otimes n_{trash}}\bigr)
  \;\approx\;
  \ket{\psi}.
\end{equation}

\section{The Quantum Metric Encoding (QME) framework}
\label{sec:method}
To address offline RL training with limited samples, we adapt the architecture of \citep{frans2024unsupervised} into a quantum framework, resulting in the Quantum Metric Encoding (QME). The QME employs a reward-supervised encoder circuit optimized for state embedding and reward prediction. Using the learned QME-embedded states and decoded rewards, we perform offline RL training. Section~\ref{sec:Quantum supervised training} provides detailed descriptions of the circuit design and training objectives, while Section~\ref{sec:Application in downstream RL tasks} demonstrates how the trained QME can be applied to downstream RL tasks, with an algorithmic flowchart presented at the end of Section~\ref{sec:Application in downstream RL tasks}.

% Motivated by evidence that trainable quantum circuits excel at metric learning \citep{lloyd2020quantum} and generalize well from limited data relative to classical models \citep{caro2022generalization}, we translate the architecture of \citep{frans2024unsupervised} into a quantum design, yielding QME. Section~\ref{sec:Quantum supervised training} details the circuit design and training objectives, and Section~\ref{sec:Application in downstream RL tasks} demonstrates how the trained QME integrates with downstream RL, with an algorithmic flowchart provided at the end of Section~\ref{sec:Application in downstream RL tasks}.

\subsection{Quantum supervised training}
\label{sec:Quantum supervised training}
Strictly, reward $r$ is determined by both $\bd{s}$ and $\bd{a}$, i.e., $r=r(\bd{s},\bd{a})$. In certain settings, one may approximate it by a state-only function, $r\approx r(\bd{s})$~\citep{frans2024unsupervised}. 
Rather than learning $r$ directly, the goal of QME is to use the reward signal $r(\cdot)$ as supervision to learn a state metric induced by a trainable quantum encoder.
% so that the quantum‑embedded state preserves reward-relevant structure and is suitable for downstream RL. 

Analogous to the quantum autoencoder (QAE) in Section~\ref{sec:PQC}, where the encoder distills key information from the original quantum state, we design our encoder-decoder circuit for the QME. Traditionally, the autoencoder aims to reconstruct the initial input at its output to effectively minimize information loss during encoding. The encoder can be learned by the difference in mutual information between two random partitions of the data \citep{tishby2000information,alemi2016deep}, while training the decoder separately \citep{frans2024unsupervised}. In quantum computing, there is no need to train the encoder and decoder separately. Quantum circuits naturally compress information while allowing for direct reward decoding. The specific circuit can be seen in detail in Fig.~\ref{fig:generalcicuit}, which mainly comprises the  following three parts: 

\begin{figure*}[th]
    \centering
    \includegraphics[width=0.85\linewidth]{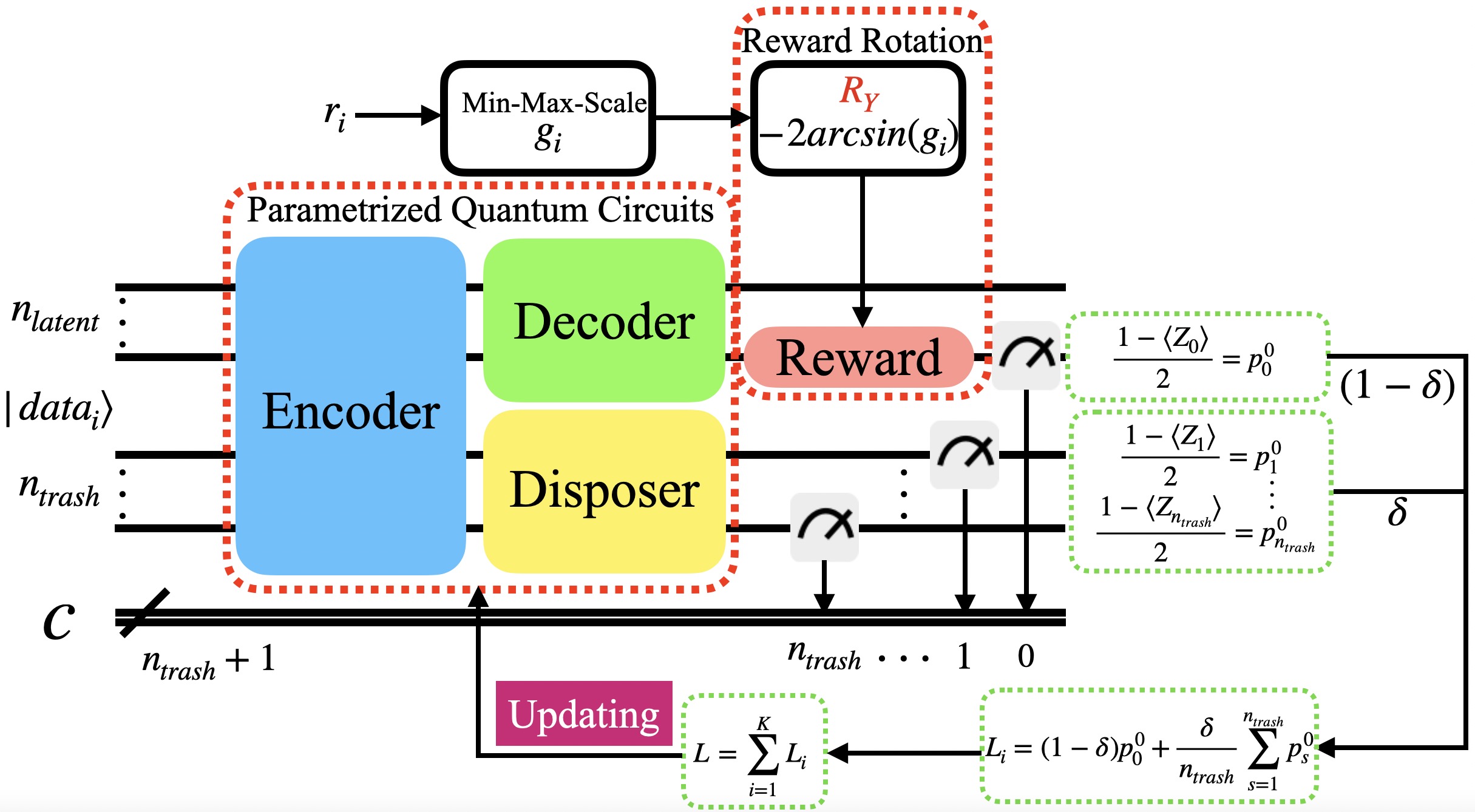}
    \caption{
    The structure and operation of the Quantum Metric Encoder (QME). Starting from quantum-encoded data, the encoder concentrates information into \(n_{\text{latent}}\) qubits and routes redundancy to \(n_{\text{trash}}\); the disposer \(U_t(\theta_t)\) resets trash to \(\ket{0}^{\otimes n_{\text{trash}}}\). A reward rotation writes the normalized reward \(g_i\) onto a target qubit; after \(U_d(\theta_d)\), an inverse rotation returns it to \(\ket{0}\) when decoded correctly. The loss (\ref{eq: li}) maximizes the \(\ket{0}\) probability for target and trash registers, with the trade-off controlled by \(\delta\).
    }

    \label{fig:generalcicuit}
\end{figure*}

\begin{itemize}
\item \textbf{Part 1 (Data Loading).}
Before applying parameterized quantum circuits (PQCs), classical data are loaded as quantum states.
We default to amplitude encoding for its logarithmic storage:
\begin{equation}
    \ket{\text{data}}=\frac{1}{\|\vec{x}\|}\sum_{i=0}^{N-1} x_i\ket{i},
\end{equation}

where \(\vec{x}\in\mathbb{R}^N\) is the input vector.
Because an \(n\)-qubit state has dimension \(2^n\), \(\vec{x}\) is zero-padded to the nearest power-of-two length as needed.

\item \textbf{Part 2 (Compression).}
An encoder \( U_e(\theta_e) \) transforms the input quantum state into a compressed representation by disentangling the relevant information into $n_{latent}$ latent qubits. The encoder compresses the quantum data into a lower-dimensional state, while the remaining irrelevant information is transferred to the $n_{trash}$ trash qubits. A parameterized trash disposer \( U_t(\theta_t) \) then maps the trash qubits to \( \ket{0}^{\otimes n_{trash}} \), removing any residual information from the original state and ensuring the compression of the quantum data.

\item \textbf{Part 3 (Reward Decoding).}
We replace the reconstruction part, which would normally involve the conjugate transpose of the encoding process in standard QAE, with direct reward decoding. For each sample \(i\) with reward \(r_i\), let \(g_i \in [0,1]\) represent the min–max normalized reward from the training set. The normalized reward \(g_i\) is encoded onto a target qubit via a rotation \(R_y(-2\arcsin(g_i))\). After passing through the decoder \(U_d(\theta_d)\), we apply this rotation so that the target qubit is driven to \( |0\rangle \) state when the reward is correctly decoded.

\end{itemize}

The target ideal quantum state evolution is thus:
\begin{align}
\ket{0} &\rightarrow 
\ket{\text{data}} (=\frac{1}{\|\vec{x}\|} \sum_{i=0}^{N-1} x_i \ket{i}), \\ &\rightarrow 
\ket{\text{compressed}}_{n_{latent}} \otimes U_{t}(\theta_{t})^\dagger \ket{0}_{n_{trash}}, \\ &\rightarrow 
\ket{\text{others}}_{n_{latent}-1} \otimes \ket{\text{scaled\, reward}} \otimes \ket{0}_{n_{trash}}, \\ &\rightarrow 
\ket{\text{others}}_{n_{latent}-1} \otimes \ket{0} \otimes \ket{0}_{n_{trash}}.
\end{align}
For the training process, considering the overall objective of our model, the goal is to have both the trash qubits and the target qubit for reward learning in the $|0\rangle$ state. We aim to maximize the sum of the probabilities of achieving the $|0\rangle$ state, which means learning to minimize the probability of the $|1\rangle$ state. A hyperparameter $\delta$ is introduced to adjust the weight of the compression information loss relative to the total loss function. The individual target $L_i$ for the $i$-th data sample in Figure~\ref{fig:generalcicuit} can be defined as:

\begin{align}
\label{eq: li}
L_i &= (1-\delta)(1 - \langle Z_0 \rangle) + \frac{\delta}{n_{\text{trash}}} \sum_{s=1}^{n_{\text{trash}}} (1 - \langle Z_s \rangle)\\
   &=\; (1-\delta)p_0^0 + \frac{\delta}{n_{\text{trash}}} \sum_{s=1}^{n_{\text{trash}}} p_s^0,
\end{align}
where 1 to $n_{trash}$ are indices of trash qubits and 0 is the index of target qubits for reward decoding. \( p_0^0 \) is the probability of the target qubit being in the state \( |0\rangle \), \( p_s^0 \) is the probability of the \( s \)-th trash qubit being in the state \( |0\rangle \).
The first term of this loss function addresses the reward learning for the target qubit. The second term corresponds to the compression of information onto the trash qubits, as described in (\ref{eq: qae}).
Assume that there are $K$ training samples in total, then the loss function $L$ for the training is:

\begin{equation}
L=\frac{1}{K}\sum_{i=1}^K L_i.
\end{equation}

In this way, we have implemented a quantum metric encoder that performs state metric and reward learning simultaneously. This ensures that the generation of the reward is based on the most relevant key information within the state given by the learned metric.

\subsection{Application in downstream Reinforcement learning (RL) tasks}
\label{sec:Application in downstream RL tasks}
After establishing the QME quantum circuits and training framework, we replace each original state with its QME-embedded counterpart for downstream RL. The embedded state is taken from the latent qubits at the bottleneck, while the decoder’s output provides a surrogate reward. In this way, a limited offline dataset $\mathcal{D}$ is transformed into a quantum-augmented dataset $\mathcal{E}$ suitable for offline RL training. The key steps are as follows:

\begin{itemize}
\item \textbf{Step 1.} 
% After training QME, for each tuple $(\bd{s},\bd{a},\bd{r},\bd{s'})\in \mathcal{D}$, we feed $\bd{s}$ and $\bd{s'}$ into the QME encoder to obtain their quantum embeddings, denoted $\bd{s_q}$ and $\bd{s'_q}$
After training QME, for each tuple $(\bd{s},\bd{a},r,\bd{s'})\in \mathcal{D}$, we feed $\bd{s}$ and $\bd{s'}$ into the QME encoder and denote their quantum embeddings -- the  QME embedded states --  as $\bd{s_q}$ and $\bd{s'_q}$. $n_{\text{latent}}$ is the number latent qubits at the bottleneck (i.e., after $U_e(\theta_e)$ and before $U_d(\theta_d)$).

\item \textbf{Step 2.} The decoded reward $r_q$ is captured from the decoder’s readout qubit before applying the reward-rotation gate. The measurement yields a probability 
$p\in[0,1]$. Using the minimum $r_{\min}:= \min_{(\bd{s},\bd{a},r,\bd{s'})\in D} r,$ and maximum $r_{\max}:= \max_{(\bd{s},\bd{a},r,\bd{s'})\in D} r$ observed from the training data $\mathcal{D}$, we map this probability back to a synthetic reward via inverse min–max scaling:
$$
r_q = r_{min}+p(r_{max}-r_{min}).
$$
\item \textbf{Step 3.} Replace each tuple with $(\bd{s_q},\bd{a},r_q,\bd{s'{_q}})$ in $\mathcal{D}$ to obtain the quantum dataset $\mathcal{E}$. Then train the model using the corresponding RL algorithm based on $\mathcal{E}$.
\end{itemize}

\begin{figure*}[h!]
    \centering
    \includegraphics[width=0.5\textwidth]{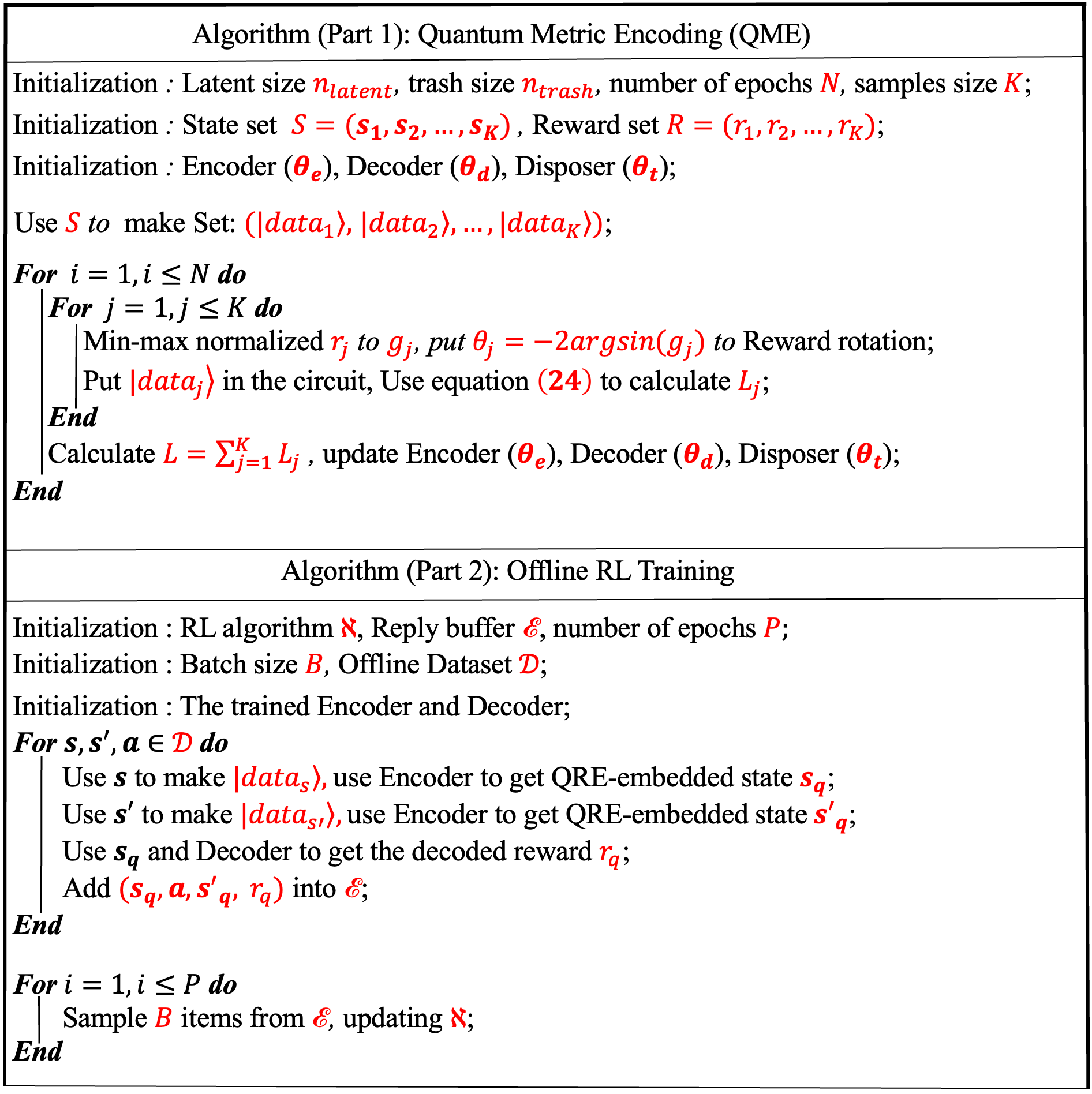}
    \caption{Algorithm 1: QME and Offline RL Training Structure. Part 1 illustrates the process of training the QME circuit, as shown in Section ~\ref{sec:Quantum supervised training}. Part 2 describes the subsequent RL training process using QME-embedded states and decoded reward, detailed in Section ~\ref{sec:Application in downstream RL tasks}.}
    \label{fig:Algorithm}
\end{figure*}

It is worth noting that the states have already been transformed into their QME-embedded representations, thereby altering their original dimensionality. Consequently, during testing, each test state must also be passed through the same QME Encoder to obtain its quantum embedding before the corresponding action can be generated. For clarity, the complete workflow of our algorithm is illustrated in Fig.~\ref{fig:Algorithm}. In the following Section~\ref{sec:experiment}, we apply our proposed method to practical RL tasks to further demonstrate its effectiveness.

\newpage
\section{Experiment}
\label{sec:experiment}

In this section, we outline the experimental protocol and configuration in Sec.~\ref{sec:experiment_protocol}, and present the main results and discussion in Sec.~\ref{sec:experiment_results}.

\subsection{Experiment Protocol and Configuration}
\label{sec:experiment_protocol}

\paragraph{\textbf{Key package settings.}} 
To ensure reproducibility, we pin the following libraries: For the quantum-training stack: \textbf{qiskit}(v.1.0.2), \textbf{qiskit-aer} (v.0.14.0.1), \textbf{qiskit{\_}algorithm} (v.0.3.0), and \textbf{qiskit{\_}machine{\_}learning} (v.0.7.2). For the RL stack: \textbf{gym} (v.0.23.1), \textbf{mujoco} (v.2.3.7), and \textbf{D4RL} (v.1.1).

\paragraph{\textbf{QME Training Settings}} 
\begin{itemize}[label=-, left=2em, labelsep=1em]
    
    \item \textbf{Dataset}: We utilize three well-known datasets provided by \textbf{D4RL}\footnote{\url{https://github.com/Farama-Foundation/D4RL}}: 
    \texttt{"bullet-hopper"}, \texttt{"bullet-halfcheetah"}, and \texttt{"bullet-ant"}. For each dataset, we draw a \textbf{fixed random subset of 100 transitions} (seed~\textbf{0}) to form a limited sample regime as a training set.
    
    \item \textbf{Parameters}: The state dimensions for \texttt{"bullet-hopper"}, \texttt{"bullet-halfcheetah"}, and \texttt{"bullet-ant"} are $\textbf{15}$, $\textbf{26}$, and $\textbf{28}$, respectively; we allocate $\textbf{4}$, $\textbf{5}$, and $\textbf{5}$ qubits for the encoder and fix $\textbf{1}$ additional “trash” qubit. 
    The compression weight $\delta$ is fixed at \textbf{$0.5$} to balance reward prediction with information compression. 
    The encoder, decoder, and trash disposer circuits are implemented using Qiskit’s \texttt{RealAmplitudes} ansatz, a layered hardware-efficient PQC that alternates single-qubit rotations $R_Y$ with a fixed entangling pattern (e.g., linear topology). The number of repetitions is set equal to the number of qubits in each block.
    QME parameters are optimized with COBYLA~\citep{powell1998direct}. In the following Fig.~\ref{fig:democircuit.png}, we illustrate an example of QME training on \texttt{"bullet-hopper"}.  In this example, we repeat the \texttt{Real Amplitudes} layer \textbf{4} times for the Encoder, \textbf{3} times for the Decoder, and \textbf{1} time for the Disposer.
    
    \begin{figure*}[h!]
        \centering
        \includegraphics[width=0.90\linewidth]
        {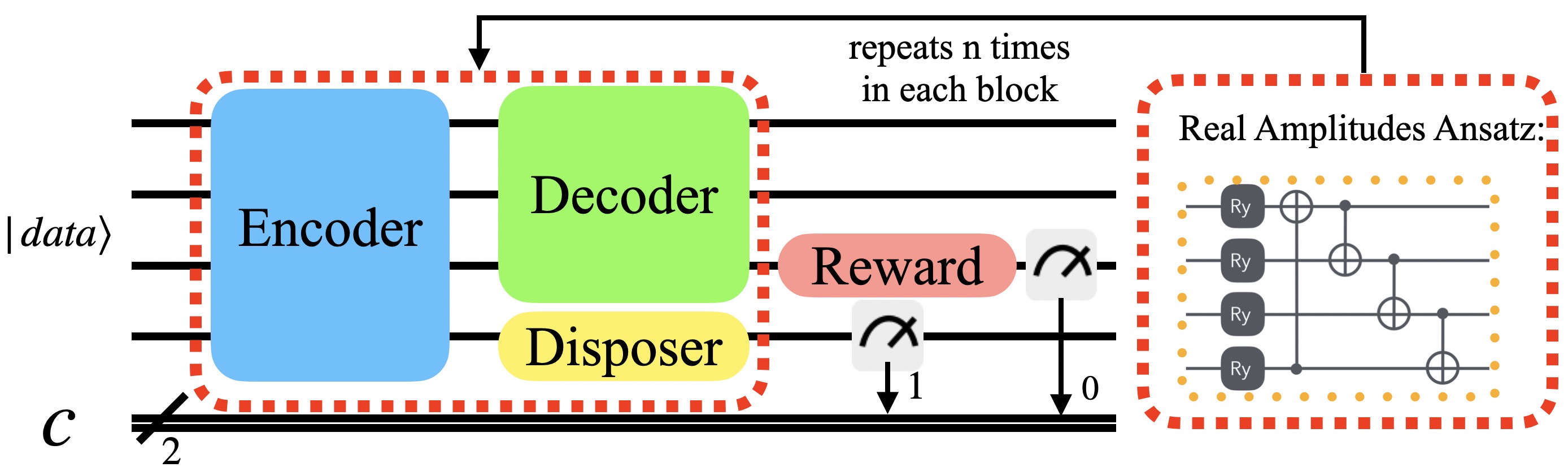}
        \caption{\textbf{QME architecture for \texttt{bullet-hopper}}. The encoder uses 4 qubits, the decoder 3 qubits, and the trash disposer 1 qubit. Each block employs qiskit’s \texttt{Real Amplitudes} ansatz, with the layer repetition set to the block’s qubit count ($n$ = 4/3/1). This configuration yields 26 trainable parameters.}
        \label{fig:democircuit.png}
    \end{figure*}

\end{itemize}

\paragraph{\textbf{RL Settings}} 

Our baselines are SAC~\citep{haarnoja2018soft2} and IQL~\citep{kostrikov2021offline}. To ensure comparability, we apply identical training controls across methods: offline batch size $=100$, epochs $=1200$, discount $\gamma=0.99$, and algorithm seed $=0$. Per-environment learning rates are fixed and shared across methods—\texttt{bullet-hopper}: $10^{-6}$; \texttt{bullet-halfcheetah}: $10^{-6}$; \texttt{bullet-ant}: $10^{-4}$. Each evaluation caps test episodes at 50 steps and is conducted every 10 epochs.
All remaining hyperparameters follow the reference implementations of SAC\footnote{\url{https://github.com/pranz24/pytorch-soft-actor-critic}} and IQL\footnote{\url{https://github.com/BY571/Implicit-Q-Learning}}. To reduce uncertainty attributable to policy stochasticity and other sources of randomness, each configuration is trained independently $5$ times. We aggregate the resulting evaluations and report the maximum overall reward, together with the mean reward at each evaluation checkpoint.

\subsection{Main Results and Discussion}
\label{sec:experiment_results}

% After training QME, we use the trained parameters to transform the training dataset into a quantum dataset and load it into the RL \textbf{ReplayBuffer}. For benchmark methods, we directly load them without transforming. 

\paragraph{\textbf{Methodology}}
We first summarize the methods used across our experiments. The only differences lie in the training datasets employed.

\begin{itemize}[label=-, left=2em, labelsep=1em]
    
    \item \textbf{Random (RD).} The dataset is collected under a uniformly random policy, serving as the lower-bound reference.
  
    \item \textbf{RL.} We train the baseline offline RL algorithms (SAC, IQL) directly on the raw original datasets, with no auxiliary preprocessing or augmentation.
  
    \item \textbf{RL + Norm.} Apply per-sample $\ell_2$ normalization to each state vector in the training data: 
    \[
    \bd{s} \leftarrow \frac{\bd{s}}{\lVert \bd{s}\rVert_2}, \qquad \bd{s'} \leftarrow \frac{\bd{s'}}{\lVert \bd{s'}\rVert_2}.
    \]
  
    \item \textbf{RL + Classical Neural Network (RL + CNN).} Using the offline tuples $(\bd{s},\bd{a},r,\bd{s'})\in \mathcal{D}$, we train a conventional feed-forward encoder–decoder to regress $r$ from $\bd{s}$. After training, we replace $r$ with the network prediction $\hat r_{C}$ for every transition, yielding $\tilde{\mathcal{D}}=\{(\bd{s},\bd{a},\hat r_{C},\bd{s'})\}$ on which SAC/IQL is trained.
  
    \item \textbf{RL + Quantum Neural Network (RL + QNN).} As above, but replacing the classical network with a quantum neural network.
  
    \item \textbf{RL + QME.} Our primary method. After training QME, follow the structure in Fig~\ref{fig:Algorithm} to make the quantum dataset $\mathcal{E}$ and train for SAC/IQL.
\end{itemize}

\paragraph{\textbf{Main results}}

\begin{itemize}[label=-, left=2em, labelsep=1em]
     \item \textbf{QME is particularly well-suited to limited data offline RL training.}
     % Motivated by the markedly low $\Delta$-hyperbolicity of QME-embedded states, we hypothesize that offline RL trained on these representations benefits substantially. 
     As an illustration with SAC, we report the mean test return (across 5 runs) smoothed with a \textbf{200}-epoch rolling window. The mean reward $R_k$ at evaluation step $k$ is
     \[
     R_k = \frac{1}{5}\sum_{i=1}^5 R_{i,k},
     \]
     where $R_{i,k}$ denotes the evaluation returns at step $k$ for run $i$. The learning curves are shown in Fig.~\ref{fig:training}.
     
     \begin{figure*}[th]
        \centering
        \includegraphics[width=\textwidth]{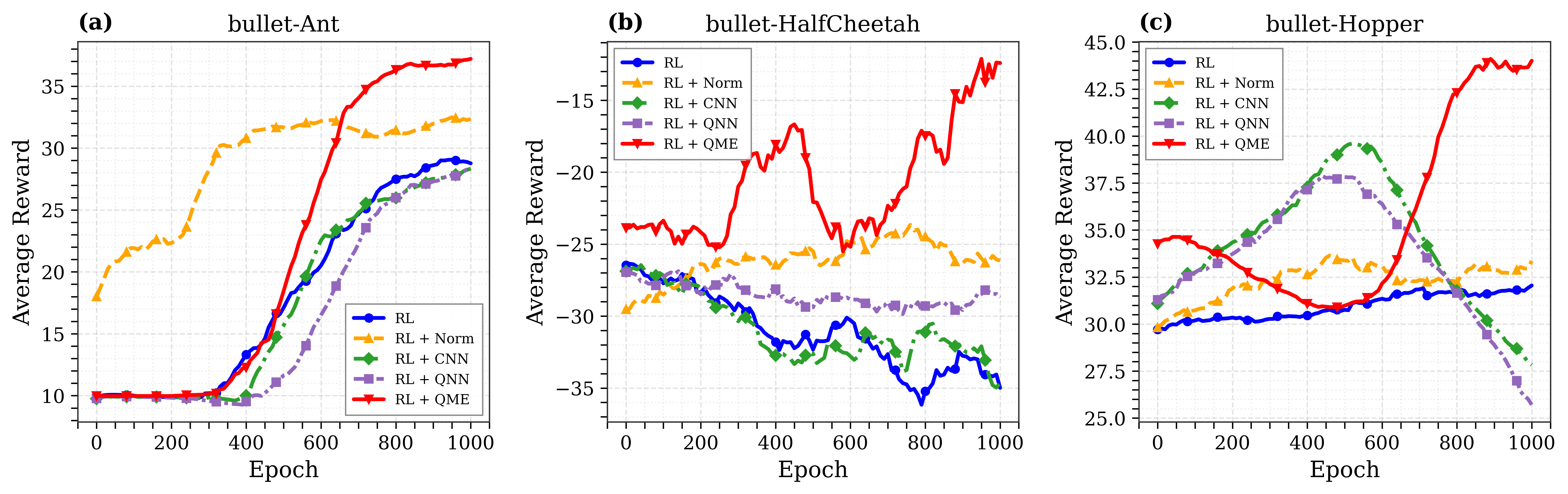}
        \caption{The average rewards during RL training for different settings across three datasets (bullet-Hopper, bullet-HalfCheetah, bullet-Ant). Under the SAC setting, the performance of ``RL + QME" (red) is markedly higher than that of ``RL + Norm" (orange) and ``RL" (blue).}
        \label{fig:training}
    \end{figure*}

    Consistent with our hypothesis, ``RL + QME" maintains better training performance across the board. By contrast, baseline RL methods overfit rapidly under limited-sample condition (Fig.~\ref{fig:training} (b)), whereas QME does not exhibit this behavior. Moreover, we find that simple state normalization (``RL + Norm") can modestly enhance the learning capacity of baseline RL; however, a substantial gap remains relative to ``RL + QME". 
    
    In addition, we report the highest reward $R_{\max}$ achieved during the whole testing phase:
    \[
    R_{\max}=\max_{i,k} R_{i,k},
    \]
    Relative to the baseline, $R_{\max}$ on ``RL + QME" is higher by \textbf{116.2\%} for SAC and \textbf{117.6\%} for IQL. Moreover, ``RL + QME" shows a statistically significant improvement over all other methods, while ``RL + Norm" performs second best. These full results are shown in Table~\ref{result_table} and Fig.~\ref{fig:result_fig}.
    
    \setlength{\tabcolsep}{20pt}
    \begin{table}[h!]
      \centering
      \caption{The complete results for SAC and IQL across the three datasets (bullet-Hopper, bullet-HalfCheetah, bullet-Ant) on different methods, the best evaluation reward achieved by each method (i.e., the maximum reward over five runs).}
      \label{result_table}
      \begin{adjustbox}{max width=\linewidth}
      \begin{threeparttable}
        \begin{tabular}{l
                        S[table-format=2.2]
                        S[table-format=2.2]
                        S[table-format=2.2]
                        S[table-format=3.2]}
          \toprule
          & \multicolumn{1}{c}{bullet-Ant}
          & \multicolumn{1}{c}{bullet-Halfcheetah}
          & \multicolumn{1}{c}{bullet-Hopper}
          & \multicolumn{1}{c}{Avg. Enh. (\%)} \\
          \cmidrule(lr){2-5}
    
          \textbf{RD} & 10.0 & -27.1 & 14.3 & \multicolumn{1}{c}{--} \\
          \midrule
    
          {SAC}       & 33.6\scriptsize\same &  -8.0\scriptsize\same & 41.5\scriptsize\same & 0.0 \\
          {+ Norm }   & 42.5\scriptsize\inc{26.5} & -3.2\scriptsize\inc{60.0} & 38.1\scriptsize\dec{8.2} & 26.1 \\
          {+ CNN}     & 31.4\scriptsize\dec{6.5} &  -8.6\scriptsize\dec{7.5} & 43.1\scriptsize\inc{3.9} &  -3.4 \\
          {+ QNN}     & 32.7\scriptsize\dec{2.7} & -19.0\scriptsize\dec{137.5} & 50.8\scriptsize\inc{22.4} & -39.2 \\
          {+ QME}     & 45.0\scriptsize\inc{33.9} &  14.6\scriptsize\inc{282.5} & 54.9\scriptsize\inc{32.3} & 116.2 \\

          \midrule
    
          {IQL}       & 41.7\scriptsize\same &  6.6\scriptsize\same  & 58.9\scriptsize\same & 0.0 \\
          {+ Norm }   & 46.0\scriptsize\inc{10.3} & 14.8\scriptsize\inc{124.2}  & 69.1\scriptsize\inc{17.3} & 50.6 \\
          {+ CNN}     & 43.5\scriptsize\inc{4.3} & -3.5\scriptsize\dec{153.0}  & 58.4\scriptsize\dec{0.8} & -49.8 \\
          {+ QNN}     & 41.1\scriptsize\dec{1.4} & -5.5\scriptsize\dec{183.3}  & 63.0\scriptsize\inc{7.0} & -59.2 \\
          {+ QME}     & 46.1\scriptsize\inc{10.6} & 27.9\scriptsize\inc{322.7}  & 70.4\scriptsize\inc{19.5} & 117.6 \\
    
          \bottomrule
        \end{tabular}
      \end{threeparttable}
      \end{adjustbox}
    \end{table}

    \begin{figure*}[th]
        \centering
        \includegraphics[width=\textwidth]{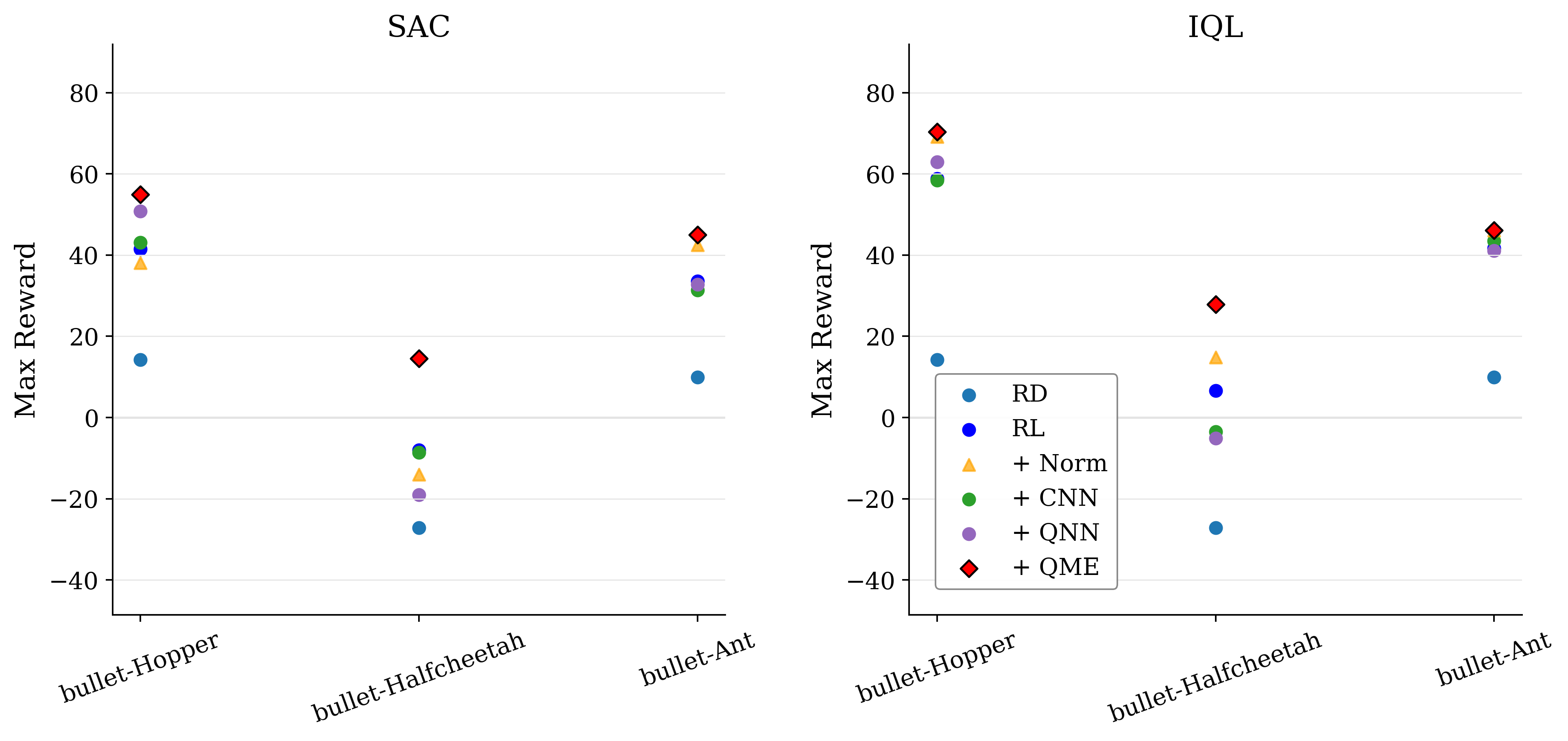}
        \caption{The complete results corresponding to Table~\ref{result_table} more intuitively highlight the advantage of ``RL + QME". }
        \label{fig:result_fig}
    \end{figure*}

    \item \textbf{QME can embed the states well into a hyperbolic space.} $\Delta$-hyperbolicity \citep{gromov1987hyperbolic} plays a pivotal role in training RL algorithms. A lower value of $\Delta$ hyperbolicity signifies a more pronounced hierarchical structure, which enhances RL performance during the testing phase \citep{cetin2022hyperbolic}. We compute the overall $\Delta$-hyperbolicity on QME-embedded states and find they exhibit substantially lower $\Delta$-hyperbolicity than the original states. Since feature normalization can artificially reduce $\Delta$-hyperbolicity, we also normalize the original states and re-evaluate; even under this controlled comparison, QME still yields markedly lower $\Delta$-hyperbolicity, the results are shown in Fig.~\ref{fig:Hyper}.
    
        \begin{figure}[h!]
          \centering
          \includegraphics[width=0.90\linewidth]{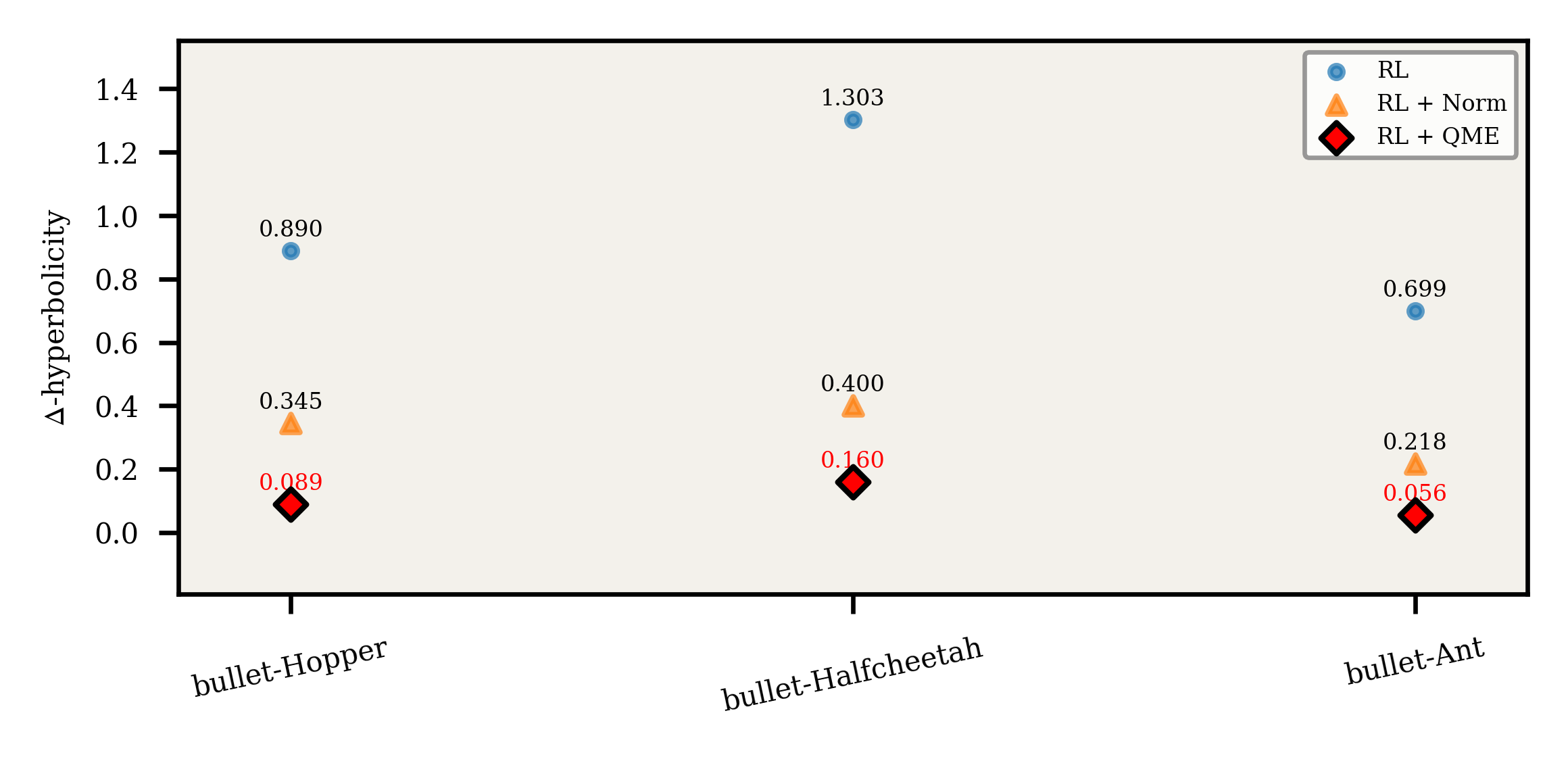} 
          \caption{The $\Delta$-hyperbolicity for different settings across three datasets (bullet-Hopper, bullet-HalfCheetah, bullet-Ant). The $\Delta$-hyperbolicity of ``RL + QME" embeddings is consistently lower than that of RL and ``RL + Norm".}
          \label{fig:Hyper}
        \end{figure}

     Taken together with the preceding conclusion, this finding suggests that the states encoded by ``RL + QME" are likely more suitable for effective learning in RL, and that the QME-embedded states are well aligned with a nearly tree-like hierarchical geometry (where $\Delta$-hyperbolicity=0 corresponds to an ideal tree hierarchy).
     For calculating $\Delta$-hyperbolicity, we follow the method of \citep{fournier2015computing}.

     This pattern mirrors the previous RL training ranking, suggesting that learning embeddings with lower $\Delta$-hyperbolicity is key to improving RL performance. 
\end{itemize}

    Overall, we observe the following experimental results:
    (1) All methods outperform the random baseline (RD), confirming the effectiveness of RL training.
    (2) Directly learning rewards through supervised methods is suboptimal, as evidenced by the inferior performance of CNN and QNN.
    (3) ``RL + Norm" provides a simple and efficient way to reduce $\Delta$-hyperbolicity and improve performance, although the improvement remains limited.
    (4) ``RL + QME" produces embedded states and decoded rewards that are more compatible with RL learning. It represents a new metric distinct from normalization. Notably, on the \texttt{bullet-HalfCheetah} task, it even reverses the sign of $R_{\max}$, demonstrating a remarkable performance gain.

    Based on these experimental results, QME demonstrates the following notable advantages:

    \textbf{(1)} QME introduces a novel metric at the quantum embedded representation level. QME embeds states in a geometry well-suited to limited data offline RL with limited samples, yielding low $\Delta$-hyperbolicity.

    \textbf{(2)} 
    Compared with CNN and QNN-based approaches, QME maintains performance under supervised reward learning and shows stronger generalization. This advantage stems from the trash qubits, enabling efficient compression and higher-level abstraction. This design supports non–end-to-end supervision and yields quantum embeddings that preserve task-relevant information for RL, consistent with \citet{frans2024unsupervised}.

    \textbf{(3)} QME attains competitive information extraction with a substantially smaller parameter budget than classical counterparts, while aligning naturally with quantum implementations. Its qubit resource requirement grows only logarithmically with the state dimension, needing $\log_2 n$ qubits for an $n$-dimensional state.

    \textbf{(4)} QME achieves this with far fewer parameters. It can mitigate overfitting in the limited data regime and improve sample efficiency relative to larger classical architectures.

    \textbf{(5)} Building on the superior performance in conventional RL tasks, QME shows potential in scenarios where the state space is high-dimensional but training samples are limited. Under such conditions, QME can still be used to approximate a reasonably effective policy, which is particularly advantageous for real-world applications and deployment.

    Therefore, the potential of QME is evident compared to traditional networks and could provide insight for the future development of combining RL with quantum and quantum-inspired methods on limited samples.

\section{Conclusion and Future Direction}
\label{sec:conclusion}

This work proposes a novel quantum-inspired encoder–decoder architecture designed for state embedding with supervised learning of reward functions. Experimental results demonstrate that under extremely limited samples, QME effectively learns the mapping from states to rewards and embeds states into representations that are well-suited for RL. This effect is further validated in offline RL training: compared with the baselines, "RL + QME" achieves an average improvement of 116.2\% on SAC and 117.6\% on IQL, while showing robustness against overfitting in limited samples regimes. We attribute this advantage to the presence of the trash qubits, which facilitates efficient information compression and abstraction. In summary, QME provides a framework that holds promise for addressing the inefficiency of offline RL caused by limited data, and it is expected to perform well in even more complex state spaces. 

QME's gains arise from learned geometry rather than quantum speedups, and the full pipeline supports fully classical simulation and parameter optimization, as used in our experiments. The computational cost of classical simulation grows with circuit width, depth, and entanglement; for high-dimensional samples or deeper ansätze, running the same circuits on quantum hardware becomes a natural alternative. In addition, when data are quantum-native (e.g., states produced by physical experiments or quantum sensors), QME can process them directly as input states, avoiding measurement and classical re-encoding overhead. We therefore view QME as a quantum-inspired module that is implementable on classical backends today and compatible with quantum processors when beneficial.\\

Nevertheless, our work still has some limitations that warrant further investigation in future studies:

\begin{itemize}
    
    \item Although our architecture can learn effectively from limited samples, whether it can function without reward information \citep{frans2024unsupervised} remains a challenging issue.

    \item We discovered this important phenomenon through experimental observation and provided a reasonable explanation. However, we lack a rigorous theoretical analysis. A more detailed analysis based on our architecture may reveal why QME can be trained efficiently on a small scale and how they improve RL training.

    \item There could exist a different classical network that can achieve results comparable to ours. However, finding such an efficient CNN poses an intriguing challenge.

    \item  We found that the $\Delta$-hyperbolicity of the QME-embedded states is very low, suggesting that hyperbolic networks, as proposed by \citep{cetin2022hyperbolic}, may be worth considering for offline RL. The quantum embedding space is an intriguing problem. Exploring its theoretical relationship with hyperbolic space is an interesting direction for future work. This could contribute to a deeper understanding on the integration of RL with quantum and quantum-inspired methods.  

\end{itemize}
Addressing these problems will further advance the understanding of the connection between RL and quantum computing.
\section*{Acknowledgements}
NL acknowledges funding from the Science and Technology Commission of Shanghai Municipality (STCSM) grant no. 24LZ1401200 (21JC1402900), NSFC grants No.12471411 and No. 12341104, the Shanghai Jiao Tong University 2030 Initiative, the Shanghai Science and Technology Innovation Action Plan (24LZ1401200) and the Fundamental Research Funds for the Central Universities.  
\bibliography{main.bib}
\end{document}

%% file: math_commands.tex
\usepackage{amsmath,amsfonts,bm}

\def\eqref#1{equation~\ref{#1}}
\def\1{\bm{1}}

\DeclareMathAlphabet{\mathsfit}{\encodingdefault}{\sfdefault}{m}{sl}
\SetMathAlphabet{\mathsfit}{bold}{\encodingdefault}{\sfdefault}{bx}{n}

\def\gA{{\mathcal{A}}}

\def\gD{{\mathcal{D}}}

\def\gS{{\mathcal{S}}}

\def\sP{{\mathbb{P}}}

\newcommand{\E}{\mathbb{E}}

\DeclareMathOperator*{\argmax}{arg\,max}

%% file: main.bib
@article{powell1998direct,
  title={Direct search algorithms for optimization calculations},
  author={Powell, Michael JD},
  journal={Acta numerica},
  volume={7},
  pages={287--336},
  year={1998},
  publisher={Cambridge University Press}
}

@article{silver2016mastering,
  title={Mastering the game of Go with deep neural networks and tree search},
  author={Silver, David and Huang, Aja and Maddison, Chris J and Guez, Arthur and Sifre, Laurent and Van Den Driessche, George and Schrittwieser, Julian and Antonoglou, Ioannis and Panneershelvam, Veda and Lanctot, Marc and others},
  journal={nature},
  volume={529},
  number={7587},
  pages={484--489},
  year={2016},
  publisher={Nature Publishing Group}
}

@inproceedings{zhu2023principled,
  title={Principled reinforcement learning with human feedback from pairwise or k-wise comparisons},
  author={Zhu, Banghua and Jordan, Michael and Jiao, Jiantao},
  booktitle={International Conference on Machine Learning},
  pages={43037--43067},
  year={2023},
  organization={PMLR}
}

@article{chen2024self,
  title={Self-play fine-tuning converts weak language models to strong language models},
  author={Chen, Zixiang and Deng, Yihe and Yuan, Huizhuo and Ji, Kaixuan and Gu, Quanquan},
  journal={arXiv preprint arXiv:2401.01335},
  year={2024}
}

@article{wang2023openchat,
  title={Openchat: Advancing open-source language models with mixed-quality data},
  author={Wang, Guan and Cheng, Sijie and Zhan, Xianyuan and Li, Xiangang and Song, Sen and Liu, Yang},
  journal={arXiv preprint arXiv:2309.11235},
  year={2023}
}

@article{ehrmann2023making,
  title={Making machine learning matter to clinicians: model actionability in medical decision-making},
  author={Ehrmann, Daniel E and Joshi, Shalmali and Goodfellow, Sebastian D and Mazwi, Mjaye L and Eytan, Danny},
  journal={NPJ Digital Medicine},
  volume={6},
  number={1},
  pages={7},
  year={2023},
  publisher={Nature Publishing Group UK London}
}

@article{otten2024does,
  title={Does reinforcement learning improve outcomes for critically ill patients? a systematic review and level-of-readiness assessment},
  author={Otten, Martijn and Jagesar, Ameet R and Dam, Tariq A and Biesheuvel, Laurens A and den Hengst, Floris and Ziesemer, Kirsten A and Thoral, Patrick J and de Grooth, Harm-Jan and Girbes, Armand RJ and Fran{\c{c}}ois-Lavet, Vincent and others},
  journal={Critical Care Medicine},
  volume={52},
  number={2},
  pages={e79--e88},
  year={2024},
  publisher={LWW}
}

@article{drudi2024reinforcement,
  title={A Reinforcement Learning Model for Optimal Treatment Strategies in Intensive Care: Assessment of the Role of Cardiorespiratory Features},
  author={Drudi, Cristian and Mollura, Maximiliano and Li-wei, H Lehman and Barbieri, Riccardo},
  journal={IEEE Open Journal of Engineering in Medicine and Biology},
  year={2024},
  publisher={IEEE}
}

@article{frans2024unsupervised,
  title={Unsupervised Zero-Shot Reinforcement Learning via Functional Reward Encodings},
  author={Frans, Kevin and Park, Seohong and Abbeel, Pieter and Levine, Sergey},
  journal={arXiv preprint arXiv:2402.17135},
  year={2024}
}

@article{tishby2000information,
  title={The information bottleneck method},
  author={Tishby, Naftali and Pereira, Fernando C and Bialek, William},
  journal={arXiv preprint physics/0004057},
  year={2000}
}

@article{alemi2016deep,
  title={Deep variational information bottleneck},
  author={Alemi, Alexander A and Fischer, Ian and Dillon, Joshua V and Murphy, Kevin},
  journal={arXiv preprint arXiv:1612.00410},
  year={2016}
}

@inproceedings{shor1994algorithms,
  title={Algorithms for quantum computation: discrete logarithms and factoring},
  author={Shor, Peter W},
  booktitle={Proceedings 35th annual symposium on foundations of computer science},
  pages={124--134},
  year={1994},
  organization={Ieee}
}

@article{grover1997quantum,
  title={Quantum mechanics helps in searching for a needle in a haystack},
  author={Grover, Lov K},
  journal={Physical review letters},
  volume={79},
  number={2},
  pages={325},
  year={1997},
  publisher={APS}
}

@ARTICLE{Cerezo2021-zg,
  title     = "Variational quantum algorithms",
  author    = "Cerezo, M and Arrasmith, Andrew and Babbush, Ryan and Benjamin,
               Simon C and Endo, Suguru and Fujii, Keisuke and McClean, Jarrod
               R and Mitarai, Kosuke and Yuan, Xiao and Cincio, Lukasz and
               Coles, Patrick J",
  journal   = "Nat. Rev. Phys.",
  publisher = "Springer Science and Business Media LLC",
  volume    =  3,
  number    =  9,
  pages     = "625--644",
  month     =  aug,
  year      =  2021,
  copyright = "https://www.springernature.com/gp/researchers/text-and-data-mining",
  language  = "en"
}

@article{havlivcek2019supervised,
  title={Supervised learning with quantum-enhanced feature spaces},
  author={Havl{\'\i}{\v{c}}ek, Vojt{\v{e}}ch and C{\'o}rcoles, Antonio D and Temme, Kristan and Harrow, Aram W and Kandala, Abhinav and Chow, Jerry M and Gambetta, Jay M},
  journal={Nature},
  volume={567},
  number={7747},
  pages={209--212},
  year={2019},
  publisher={Nature Publishing Group}
}

@article{Perez-Salinas2020-cp,
  title     = "Data re-uploading for a universal quantum classifier",
  author    = "P{\'e}rez-Salinas, Adri{\'a}n and Cervera-Lierta, Alba and
               Gil-Fuster, Elies and Latorre, Jos{\'e} I",
  journal   = "Quantum",
  publisher = "Verein zur Forderung des Open Access Publizierens in den
               Quantenwissenschaften",
  volume    =  4,
  number    =  226,
  pages     = "226",
  month     =  feb,
  year      =  2020,
  language  = "en"
}

@article{schuld2021effect,
  title={Effect of data encoding on the expressive power of variational quantum-machine-learning models},
  author={Schuld, Maria and Sweke, Ryan and Meyer, Johannes Jakob},
  journal={Physical Review A},
  volume={103},
  number={3},
  pages={032430},
  year={2021},
  publisher={APS}
}

@article{biamonte2017quantum,
  title={Quantum machine learning},
  author={Biamonte, Jacob and Wittek, Peter and Pancotti, Nicola and Rebentrost, Patrick and Wiebe, Nathan and Lloyd, Seth},
  journal={Nature},
  volume={549},
  number={7671},
  pages={195--202},
  year={2017},
  publisher={Nature Publishing Group UK London}
}

@article{romero2017quantum,
  title={Quantum autoencoders for efficient compression of quantum data},
  author={Romero, Jonathan and Olson, Jonathan P and Aspuru-Guzik, Alan},
  journal={Quantum Science and Technology},
  volume={2},
  number={4},
  pages={045001},
  year={2017},
  publisher={IOP Publishing}
}

@article{kumar2020conservative,
  title={Conservative q-learning for offline reinforcement learning},
  author={Kumar, Aviral and Zhou, Aurick and Tucker, George and Levine, Sergey},
  journal={Advances in Neural Information Processing Systems},
  volume={33},
  pages={1179--1191},
  year={2020}
}

@article{haarnoja2018soft,
  title={Soft actor-critic algorithms and applications},
  author={Haarnoja, Tuomas and Zhou, Aurick and Hartikainen, Kristian and Tucker, George and Ha, Sehoon and Tan, Jie and Kumar, Vikash and Zhu, Henry and Gupta, Abhishek and Abbeel, Pieter and others},
  journal={arXiv:1812.05905},
  year={2018}
}

@inproceedings{haarnoja2018soft2,
  title={Soft actor-critic: Off-policy maximum entropy deep reinforcement learning with a stochastic actor},
  author={Haarnoja, Tuomas and Zhou, Aurick and Abbeel, Pieter and Levine, Sergey},
  booktitle={International conference on machine learning},
  pages={1861--1870},
  year={2018},
  organization={PMLR}
}

@article{mnih2013playing,
  title={Playing atari with deep reinforcement learning},
  author={Mnih, Volodymyr and Kavukcuoglu, Koray and Silver, David and Graves, Alex and Antonoglou, Ioannis and Wierstra, Daan and Riedmiller, Martin},
  journal={arXiv:1312.5602},
  year={2013}
}

@article{konda1999actor,
  title={Actor-critic algorithms},
  author={Konda, Vijay and Tsitsiklis, John},
  journal={Advances in neural information processing systems},
  volume={12},
  year={1999}
}

@article{kostrikov2021offline,
  title={Offline reinforcement learning with implicit q-learning},
  author={Kostrikov, Ilya and Nair, Ashvin and Levine, Sergey},
  journal={arXiv:2110.06169},
  year={2021}
}

@book{bertsekas2014constrained,
  title={Constrained optimization and Lagrange multiplier methods},
  author={Bertsekas, Dimitri P},
  year={2014},
  publisher={Academic press}
}

@article{qi2023adaptive,
  title={An adaptive reinforcement learning-based multimodal data fusion framework for human--robot confrontation gaming},
  author={Qi, Wen and Fan, Haoyu and Karimi, Hamid Reza and Su, Hang},
  journal={Neural Networks},
  volume={164},
  pages={489--496},
  year={2023},
  publisher={Elsevier}
}

@article{liu2022avoiding,
  title={Avoiding Overfitting to the Importance Weights in Offline Policy Optimization},
  author={Liu, Yao and Brunskill, Emma},
  journal = {},
  year={2022}
}

@article{nie2022data,
  title={Data-efficient pipeline for offline reinforcement learning with limited data},
  author={Nie, Allen and Flet-Berliac, Yannis and Jordan, Deon and Steenbergen, William and Brunskill, Emma},
  journal={Advances in Neural Information Processing Systems},
  volume={35},
  pages={14810--14823},
  year={2022}
}

@article{abbas2021power,
  title={The power of quantum neural networks},
  author={Abbas, Amira and Sutter, David and Zoufal, Christa and Lucchi, Aur{\'e}lien and Figalli, Alessio and Woerner, Stefan},
  journal={Nature Computational Science},
  volume={1},
  number={6},
  pages={403--409},
  year={2021},
  publisher={Nature Publishing Group US New York}
}

@inproceedings{macaluso2024small,
  title={Small dataset, big gains: Enhancing reinforcement learning by offline pre-training with model-based augmentation},
  author={Macaluso, Girolamo and Sestini, Alessandro and Bagdanov, Andrew D},
  booktitle={Computer Sciences \& Mathematics Forum},
  volume={9},
  number={1},
  pages={4},
  year={2024},
  organization={MDPI}
}

@article{cetin2022hyperbolic,
  title={Hyperbolic deep reinforcement learning},
  author={Cetin, Edoardo and Chamberlain, Benjamin and Bronstein, Michael and Hunt, Jonathan J},
  journal={arXiv preprint arXiv:2210.01542},
  year={2022}
}

@article{lloyd2020quantum,
  title={Quantum embeddings for machine learning},
  author={Lloyd, Seth and Schuld, Maria and Ijaz, Aroosa and Izaac, Josh and Killoran, Nathan},
  journal={arXiv preprint arXiv:2001.03622},
  year={2020}
}

@article{schuld2019quantum,
  title={Quantum machine learning in feature Hilbert spaces},
  author={Schuld, Maria and Killoran, Nathan},
  journal={Physical review letters},
  volume={122},
  number={4},
  pages={040504},
  year={2019},
  publisher={APS}
}

@article{caro2022generalization,
  title={Generalization in quantum machine learning from few training data},
  author={Caro, Matthias C and Huang, Hsin-Yuan and Cerezo, Marco and Sharma, Kunal and Sornborger, Andrew and Cincio, Lukasz and Coles, Patrick J},
  journal={Nature communications},
  volume={13},
  number={1},
  pages={4919},
  year={2022},
  publisher={Nature Publishing Group UK London}
}

@incollection{gromov1987hyperbolic,
  title={Hyperbolic groups},
  author={Gromov, Mikhael},
  booktitle={Essays in group theory},
  pages={75--263},
  year={1987},
  publisher={Springer}
}

@article{fournier2015computing,
  title={Computing the Gromov hyperbolicity of a discrete metric space},
  author={Fournier, Herv{\'e} and Ismail, Anas and Vigneron, Antoine},
  journal={Information Processing Letters},
  volume={115},
  number={6-8},
  pages={576--579},
  year={2015},
  publisher={Elsevier}
}

@article{guo2025deepseek,
  title={Deepseek-r1: Incentivizing reasoning capability in llms via reinforcement learning},
  author={Guo, Daya and Yang, Dejian and Zhang, Haowei and Song, Junxiao and Zhang, Ruoyu and Xu, Runxin and Zhu, Qihao and Ma, Shirong and Wang, Peiyi and Bi, Xiao and others},
  journal={arXiv preprint arXiv:2501.12948},
  year={2025}
}

@article{peng2019advantage,
  title={Advantage-weighted regression: Simple and scalable off-policy reinforcement learning},
  author={Peng, Xue Bin and Kumar, Aviral and Zhang, Grace and Levine, Sergey},
  journal={arXiv preprint arXiv:1910.00177},
  year={2019}
}

@article{levine2018reinforcement,
  title={Reinforcement learning and control as probabilistic inference: Tutorial and review},
  author={Levine, Sergey},
  journal={arXiv preprint arXiv:1805.00909},
  year={2018}
}

@article{lillicrap2015continuous,
  title={Continuous control with deep reinforcement learning},
  author={Lillicrap, Timothy P and Hunt, Jonathan J and Pritzel, Alexander and Heess, Nicolas and Erez, Tom and Tassa, Yuval and Silver, David and Wierstra, Daan},
  journal={arXiv preprint arXiv:1509.02971},
  year={2015}
}
